\documentclass[letterpaper, 10 pt, conference]{ieeeconf}  % Comment this line out if you need a4paper

\IEEEoverridecommandlockouts                              % This command is only needed if 
\usepackage{graphicx} % for pdf, bitmapped graphics files
\graphicspath{{./images/}}
\usepackage{subfigure}
\usepackage{epsfig} % for postscript graphics files
\usepackage{mathptmx} % assumes new font selection scheme installed
\usepackage{times} % assumes new font selection scheme installed
\usepackage{amsmath} % assumes amsmath package installed
\usepackage{amssymb}  % assumes amsmath package installed
\usepackage[utf8]{inputenc}
\usepackage{bbold}
\usepackage[strings]{underscore}
\usepackage{cite}
\usepackage{xcolor}

\title{\LARGE \bf
Clustering Trust Dynamics in a Human-Robot\\ Sequential Decision-Making Task
}

\author{Shreyas Bhat$^1$, Joseph B. Lyons$^2$, Cong Shi$^1$, and X. Jessie Yang$^1$
\thanks{*This work was supported in part by the Air Force Office of Scientific Research (Grant \#FA9550-20-1-0406).}% <-this % stops a space
\thanks{$^{1}$Shreyas Bhat, Cong Shi, and X. Jessie Yang are with the Industrial and Operations Engineering Department, University of Michigan, Ann Arbor, MI 48109
      {\small(email: shreyasb@umich.edu, shicong@umich.edu, xijyang@umich.edu) }}%
\thanks{$^{2}$Joseph B. Lyons is with the Air Force Research Laboratory
      {\small(email: joseph.lyons.6@us.af.mil)}}%
}

\begin{document}
\maketitle
\thispagestyle{empty}
\pagestyle{empty}
%%%%%%%%%%%%%%%%%%%%%%%%%%%%%%%%%%%%%%%%%%%%%%%%%%%%%%%%%%%%%%%%%%%%%%%%%%%%%%%%
\begin{abstract}
In this paper, we present a framework for trust-aware sequential decision-making in a human-robot team. We model the problem as a finite-horizon Markov Decision Process with a reward-based performance metric, allowing the robotic agent to make trust-aware recommendations. Results of a human-subject experiment show that the proposed trust update model is able to accurately capture the human agent's moment-to-moment trust changes. Moreover, we cluster the participants' trust dynamics into three categories, namely, Bayesian decision makers, oscillators, and disbelievers, and identify personal characteristics that could be used to predict which type of trust dynamics a person will belong to. We find that the disbelievers are less extroverted, less agreeable, and have lower expectations toward the robotic agent, compared to the Bayesian decision makers and oscillators. The oscillators are significantly more frustrated than the Bayesian decision makers.
% We report some significant individual differences between these clusters through one-way ANOVA on the collected data.

\end{abstract}

% \begin{keywords}
% Acceptability and trust, human-robot teaming, human-robot collaboration, planning under uncertainty, trust dynamics, individual differences
% \end{keywords}

%%%%%%%%%%%%%%%%%%%%%%%%%%%%%%%%%%%%%%%%%%%%%%%%%%%%%%%%%%%%%%%%%%%%%%%%%%%%%%%%
\section{INTRODUCTION}
\noindent Trust has been identified as a key factor for effective human-robot interaction. Consequently, substantial research efforts have been devoted to identifying factors that influence humans' trust in robots \cite{Hancock2020}, developing computational models for trust estimation \cite{Optimo, BetaDistPaper}, and developing trust-aware robots \cite{trustPomdpLong, ReversePsychologyModel}. 

Previous computational models of trust\cite{Optimo, BetaDistPaper} require binary performance measures of the autonomy. However, such measures cannot be directly applied in scenarios wherein the robotic agent needs to perform complex trade off decisions to maximize the cumulative reward and hence the autonomy performance is more difficult to quantify. In addition, although previous studies have revealed the existence of different types of trust dynamics \cite{BetaDistPaper, ClusteringTrustDynamics}, no research has tried to associate personal characteristics with the type of trust dynamics.

To fill these research gaps, we develop a novel reward-based performance metric to drive the trust estimation algorithm in a Markov Decision Process (MDP). The proposed performance metric is a binary surrogate of the autonomous agent's reward function, which allows our autonomous agent to utilize the trust update model developed in \cite{BetaDistPaper} for trust-aware decision-making. Moreover, from the data collected through human-subject experiments, we analyze how human trust evolves with their earned reward over repeated interactions with the autonomy. We find three distinct types of trust dynamics through $k-$means clustering analysis and examine associations between personal characteristics and type of trust dynamics.

\section{RELATED WORK}
\label{sec:relatedWork}
% In recent years, a substantial amount of research has focused on the assessment of trust between humans and machines. Efforts are also being made to use trust to drive autonomous agents' behavior. In this section, we detail some of the work that has been done in this area of research.
% \subsection{Modeling (Snapshot) Trust in Automation and Trust Dynamics}
Extensive work has been done in identifying factors that affect trust in automation. However, most of these studies provide a snapshot view of trust, and failed to characterize the dynamics of trust \cite{Yang2021}. More recently, dynamic models of trust has been proposed, including the auto-regressive moving average vector (ARMAV) model \cite{armav-trust-model}, the Online Probabilistic Trust Inference Model (OPTIMo) model \cite{Optimo}, and the experience-based model \cite{BetaDistPaper}. These models enable a robot to estimate a human's trust in real time and has led to the development of trust-aware robots, allowing robots to reason upon human trust in their planning \cite{trustSeekingRobots, trustPomdpLong, trustWorkloadPomdpII, ReversePsychologyModel}.

Most existing research investigating individual differences aims to find associations between (snapshot) trust and individual characteristics, including propensity to trust automation \cite{merritt-ilgen-propensity}, personality trait of neuroticism \cite{Szalma2011IndividualDI}, and dispositional expectations of technology \cite{MerrittPASscale, LyonsGuznovPAS}. It is possible that these individual differences will be differentially represented within trust profiles related to trust dynamics. Recent research \cite{BetaDistPaper} reveal the existence of different types of trust dynamics, namely, rational decision maker, oscillator, and disbeliever. It is reasonable to investigate the association between a human agent's trust dynamics and factors that are shown to affect a human agent's (snapshot) trust.

% Most existing research investigating individual differences aims to find associations between individual characteristics and (snapshot) trust. Dispositional trust represents an individual's domain-independent "overall" tendency to trust automation. Notably, one's propensity to trust automation is target-agnostic and thus treats trust as a form trait in and of itself. Such measures are useful in tapping into one's general tendency to trust. Individuals with high propensity to trust automation had a higher difference in post-task trust between a reliable and faulty automation \cite{merritt-ilgen-propensity}. Yet, beyond one's propensity to trust, there are individual differences that can influence one's attitudes toward automation. The personality trait of neuroticism was found to be negatively correlated with agreement with automation in \cite{Szalma2011IndividualDI}. The Perfect Automation Schema which represents one's dispositional expectations of technology \cite{MerrittPASscale} is a significant factor in trust in automation as it it has been associated with state trust in automation \cite{LyonsGuznovPAS}. It is possible that these individual differences will be differentially represented within trust profiles related to trust dynamics.  

% Recent research \cite{BetaDistPaper} reveal the existence of different types of trust dynamics, namely, rational decision maker, oscillator, and disbeliever. It is reasonable to investigate the association between a human agent's trust dynamics and factors that are shown to affect a human agent's (snapshot) trust. 

\section{PROBLEM FORMULATION}
\label{sec:model}
In this section, we propose a finite horizon Markov Decision Process (MDP) for modeling and incorporating trust in the decision-making system of a robotic agent. The robot provides recommendations to its human partner about the action that they should take, but the final decision lies with the human. The specific scenario we target is an 'Intelligence, Surveillance, and Reconnaissance' (ISR) mission in which a human soldier teams up with an \emph{intelligent} drone to search through a town for threats. The drone guides the soldier on whether s/he should breach a site directly or deploy a Robotic Armored Rescue Vehicle (RARV). Using the RARV prevents any health loss to the soldier in the presence of a threat but takes additional time to the team; on the contrary, breaching a site directly is faster, but the soldier will be harmed if s/he encounters a threat. Here, two natural (but conflicting) goals that arise are to minimize any damage to the soldier and minimize the time to search through all the sites. 

The key components of the MDP model of the ISR mission are given below.

\subsubsection{\bf States}
We use the estimated trust of the human on the robot as the state. Specifically, we use the trust model in \cite{BetaDistPaper}, wherein trust level at the $i^\text{th}$ stage, $t_i$, follows a Beta distribution, i.e.,
$
t_i \sim Beta(\alpha_i, \beta_i),
$
where $\alpha_i$ and $\beta_i$ are the positive experience and negative experience at time $i$. And, the state at stage $i$ is specified by the tuple $(\alpha_i, \beta_i)$. 

\subsubsection{\bf Actions}
The drone has two actions: Recommend to use or not use the RARV.

\subsubsection{\bf Human behavior model}
We assume that human will accept the recommendation given by the robotic agent with probability $t_i$; and the human chooses the opposite action of the one that was recommended with probability $1- t_i$.

% \begin{align}
% \label{eq::Behavior}
% \begin{split}
% P(a_h=a_r) &= t_i,\\
% P(a_h=1-a_r) &= 1-t_i.    
% \end{split}
% \end{align}
% Here $t_i$ is the trust level at the $i^\text{th}$ search site.

\subsubsection{\bf Reward function}
We use the weighted sum of the health loss cost and the time cost to define the cost of choosing an action. Similar to \cite{ReversePsychologyModel}, we add a trust-gaining term to incentivize the autonomous agent to gain trust. As a result, the reward $R_i$ at the $i^\text{th}$ stage is defined as
% seek low-trust  exploits this human behavior by always recommending the opposite action. Trust being low, the human chooses the action opposite to the recommendation, thus increasing performance. We call this behavior of the autonomy as \emph{reverse-psychology}. Since such deceptive behavior is undesirable, we add a trust gain reward to the reward function to incentivize the autonomous agent to make righteous recommendations.

% if we only use this \emph{performance} based reward, the autonomous agent learns that for most people, trust decreases after failures more easily than increasing after successes \cite{BetaDistPaper}, and thus it exploits this human behavior by always recommending the opposite action. Trust being low, the human chooses the action opposite to the recommendation, thus increasing performance. We call this behavior of the autonomy as \emph{reverse-psychology}. Since such deceptive behavior is undesirable, we add a trust gain reward to the reward function to incentivize the autonomous agent to make righteous recommendations.
\begin{equation}
\label{eq::reward-function}
R_i(a) = -w_h h(a) - w_c c(a) + \gamma_i\cdot \mathbb{1}(A),
\end{equation}
where $w_h$ and $w_c$ are the weights for the health loss cost and time loss cost respectively. We define $A$ as the event when trust increases, i.e., $\mathbb{1}(A) = 1$ if the performance of the autonomous agent is a success and $\mathbb{1}(A) = 0$ otherwise. $\gamma_i (=w_t \sqrt{N-i})$ is a weight given to the trust gain reward that decreases with the stage number. The idea behind this is to support trust-gaining behavior near the current stage, and task performance optimizing behavior towards the later stages of planning. 
% In particular, we set $\gamma_i = $.

\subsubsection{\bf Transition function}
\label{sec::TransitionFunction}
The state $(\alpha_i, \beta_i)$ is updated by
\begin{equation}
\label{eq::alpha-update}
( \alpha _{i} ,\beta _{i}) =\begin{cases}
\left( \alpha _{i-1} +w^{s} ,\beta _{i-1}\right) & \text{if } p_{i} =1,\\
\left( \alpha _{i-1} ,\beta _{i-1} +w^{f}\right) & \text{if } p_{i} =0.
\end{cases}
\end{equation}
Here, $p_{i}$ is the autonomous agent's performance at the $i^\text{th}$ stage. 
% As the autonomous agent's performance is not binary, i.e., the autonomous agent is either correct or wrong, the definition of $p_i$ in \cite{BetaDistPaper} cannot be directly applied. Rather, 
Since the team's goal is to maximize earned reward, we define $p_i=1$ if at the $i^\text{th}$ site, the immediate reward for following the recommendation by the drone was greater than that for not following the recommendation and $p_i=0$ otherwise.

% As per \cite{BetaDistPaper}, w
We fit the parameters $(\alpha_0, \beta_0, w^s, w^f)$ for each participant to model their trust dynamics. We use gradient descent to estimate the parameters in real time after each feedback from the participant. 
% We use the digamma function approximation presented in \cite{digammaApproximation} to approximate the gradients of the beta distribution function. 
We use value iteration to solve the MDP.

\section{Experiment}
\label{sec:case}
This section describes details of the human-subject experiment. The experiment complied with the American Psychological Association code of ethics and was approved by the Institutional Review Board at the University of Michigan.

\subsection{Participants}
A total of 46 adults participated in the study. One participant's data was discarded as the participant failed the attention check. The remaining 45 participants consisted of 21 females and 24 males (Age: Mean = 22.8 years, \textit{SD} = 3.6 years). Each participant was reimbursed with a base pay of \$20 with a bonus of up to \$10 based on their performance, which was measured by the time taken by the participants to complete the task and the final health level of the soldier.

\subsection{Testbed}
\begin{figure}[h]
\centering
\subfigure[A view of the testbed]{
\includegraphics[width=0.46\linewidth]{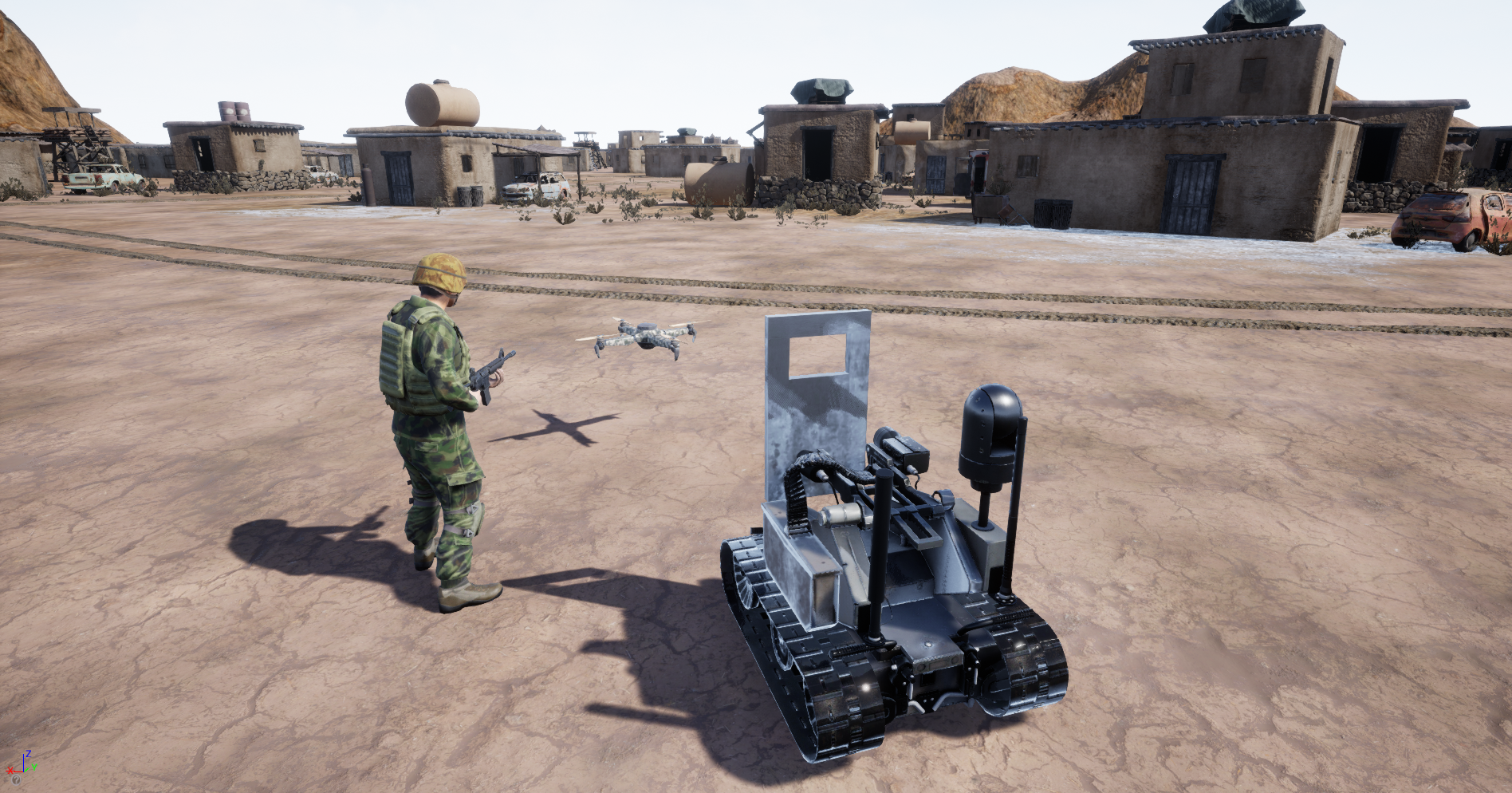}
\label{fig::Background}}
\subfigure[An example of the GUI]{
\includegraphics[width=0.46\linewidth]{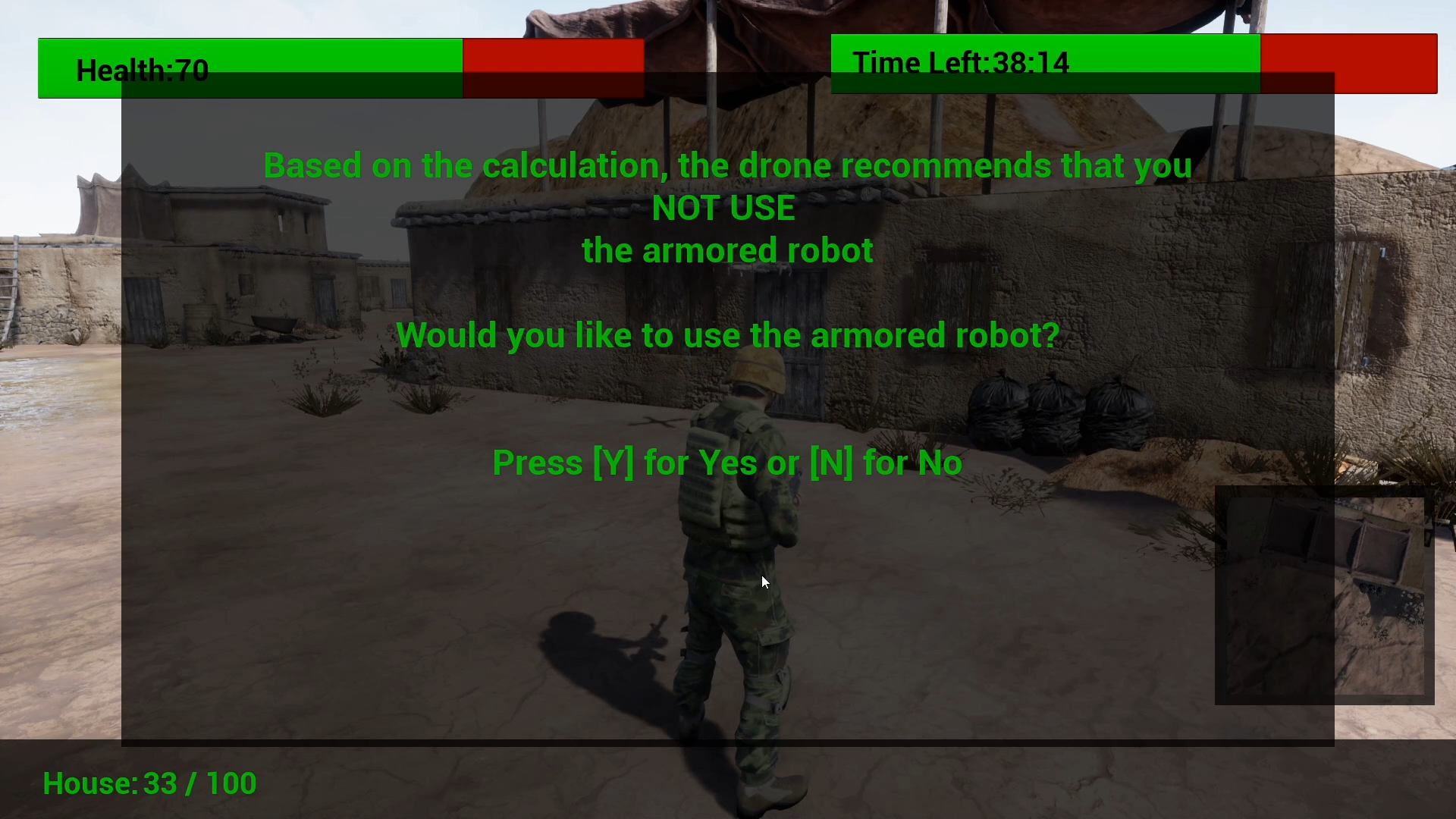}
\label{fig::Recommendation}}
\caption{The testbed developed in the Unreal Engine 4 game engine}
\label{fig::Testbed}
\end{figure}

\begin{figure}[h]
\centering
\subfigure[No Threat, RARV Not Used]{
\includegraphics[width=0.46\linewidth]{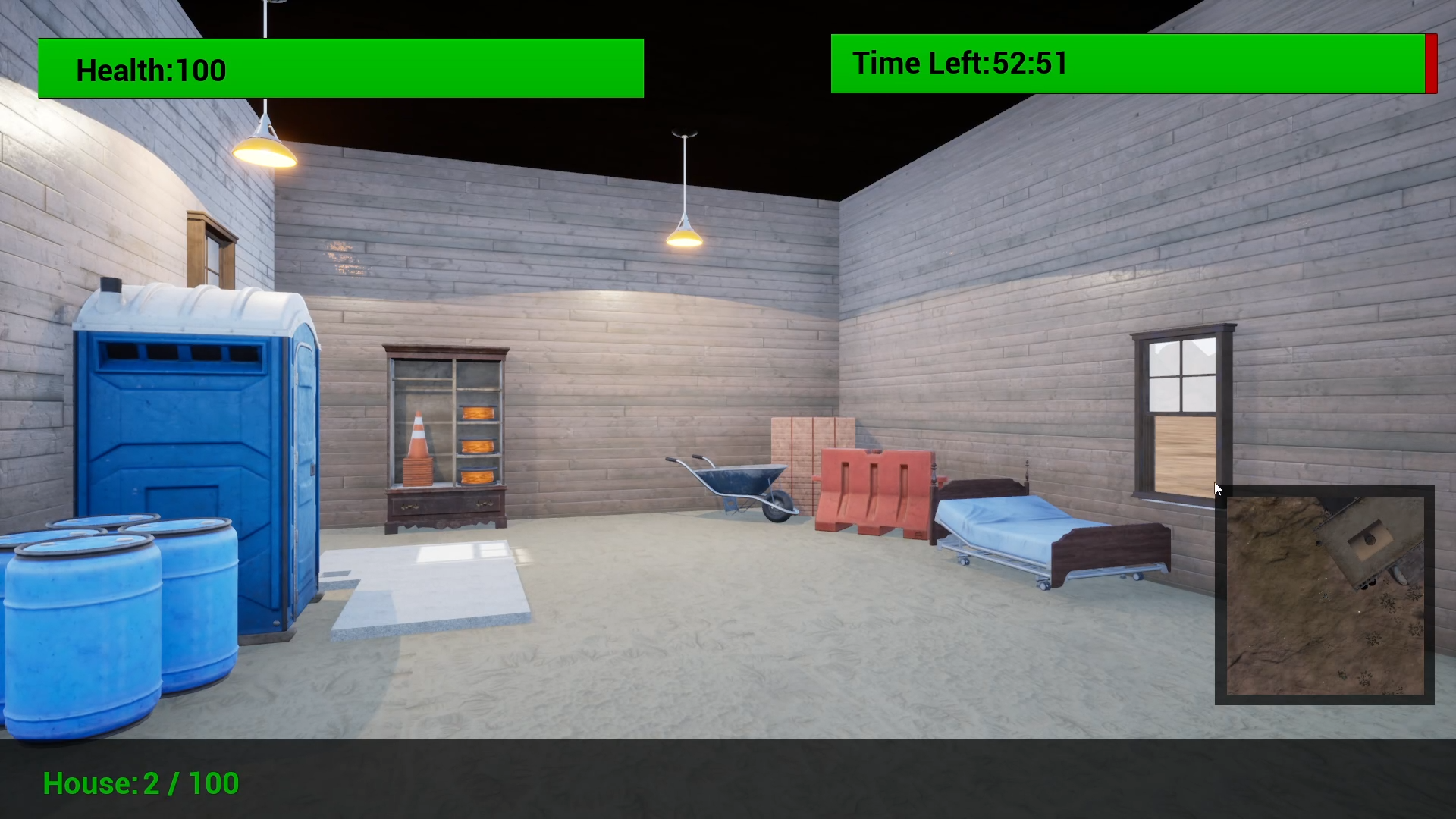}
\label{fig::Outcomes1}}
\hfil
\subfigure[No Threat, RARV Used]{
\includegraphics[width=0.46\linewidth]{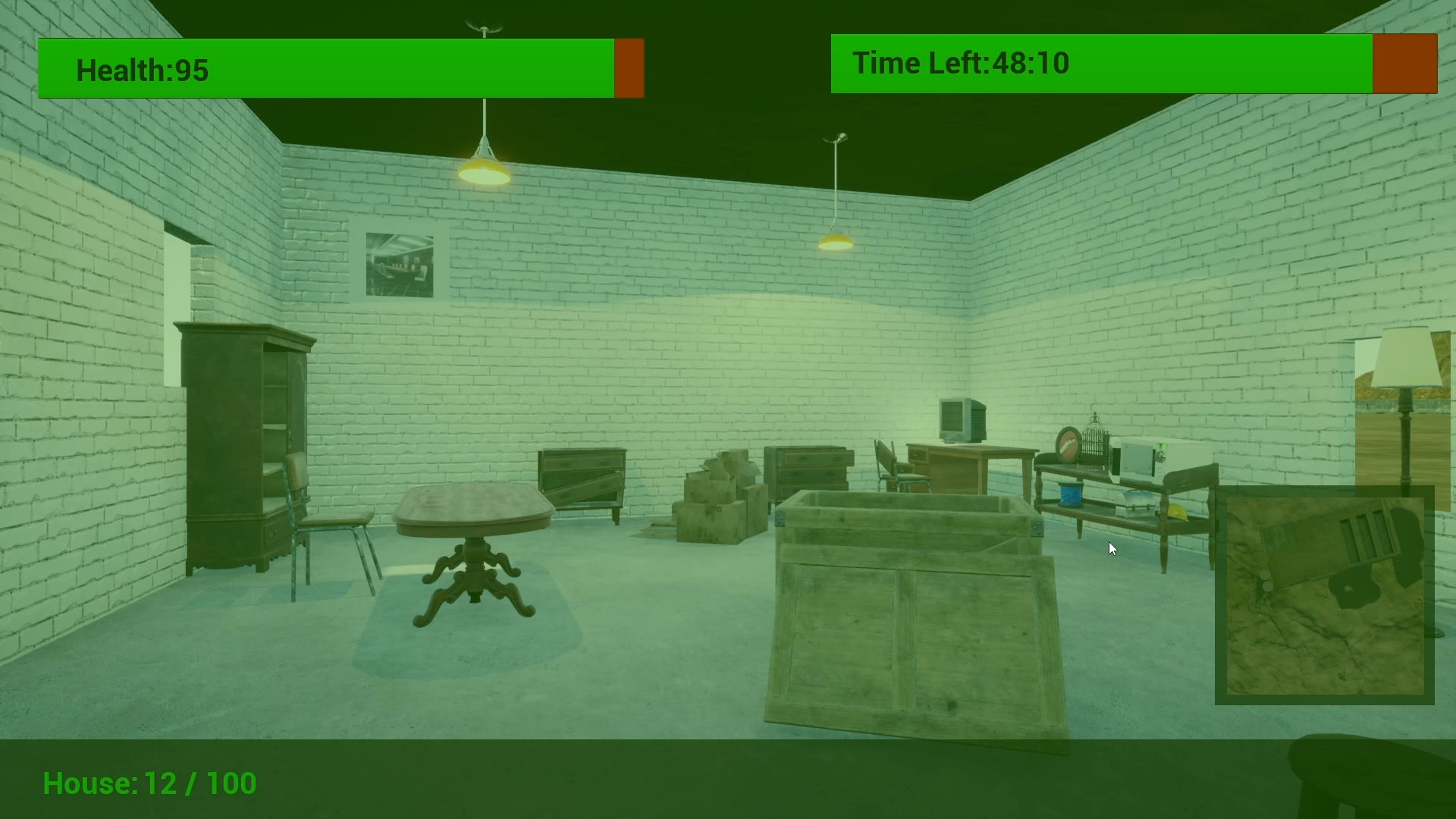}
\label{fig::Outcomes2}}
\subfigure[Threat, RARV Not Used]{
\includegraphics[width=0.46\linewidth]{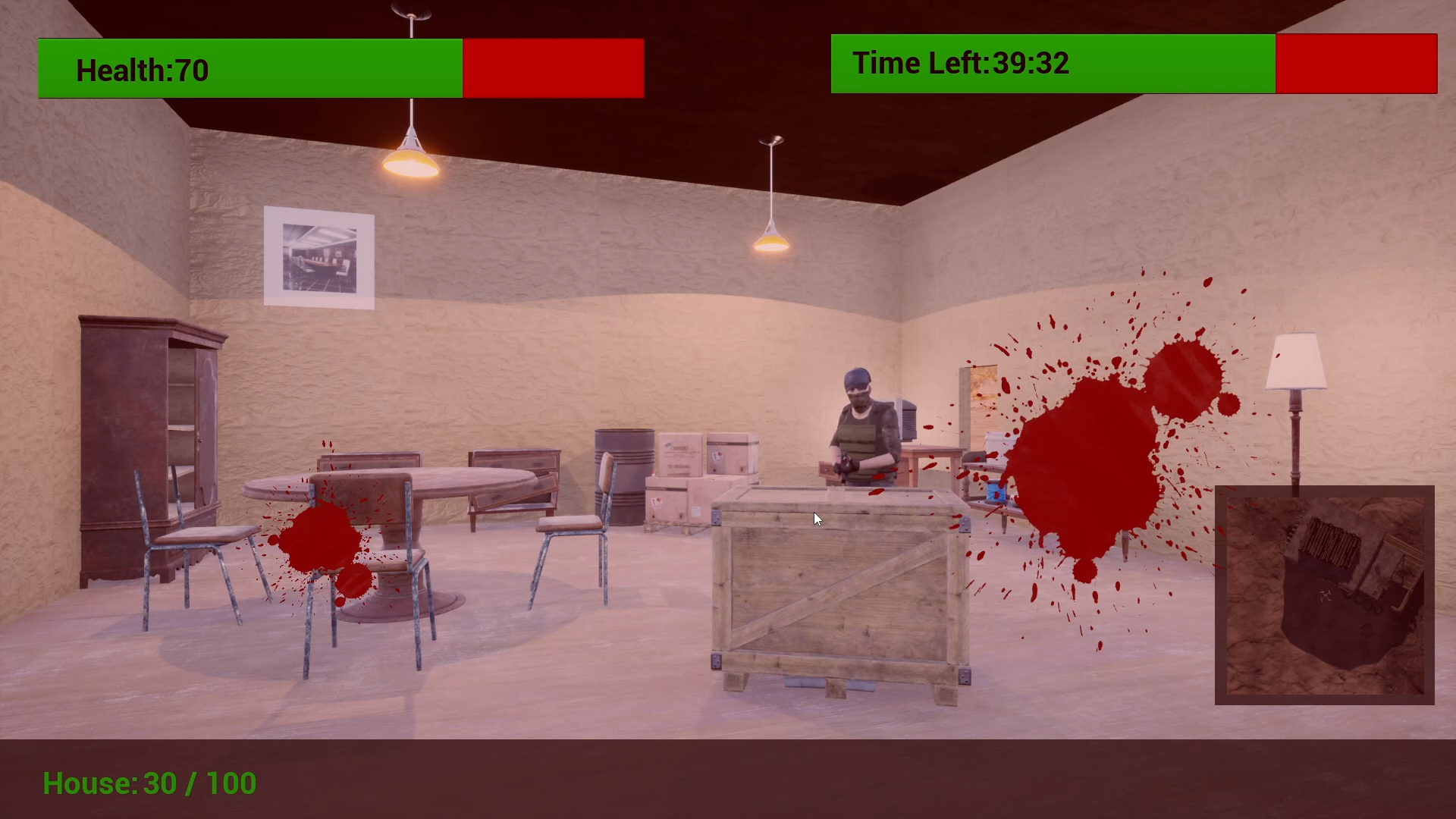}
\label{fig::Outcomes3}}
\hfil
\subfigure[Threat, RARV Used]{
\includegraphics[width=0.46\linewidth]{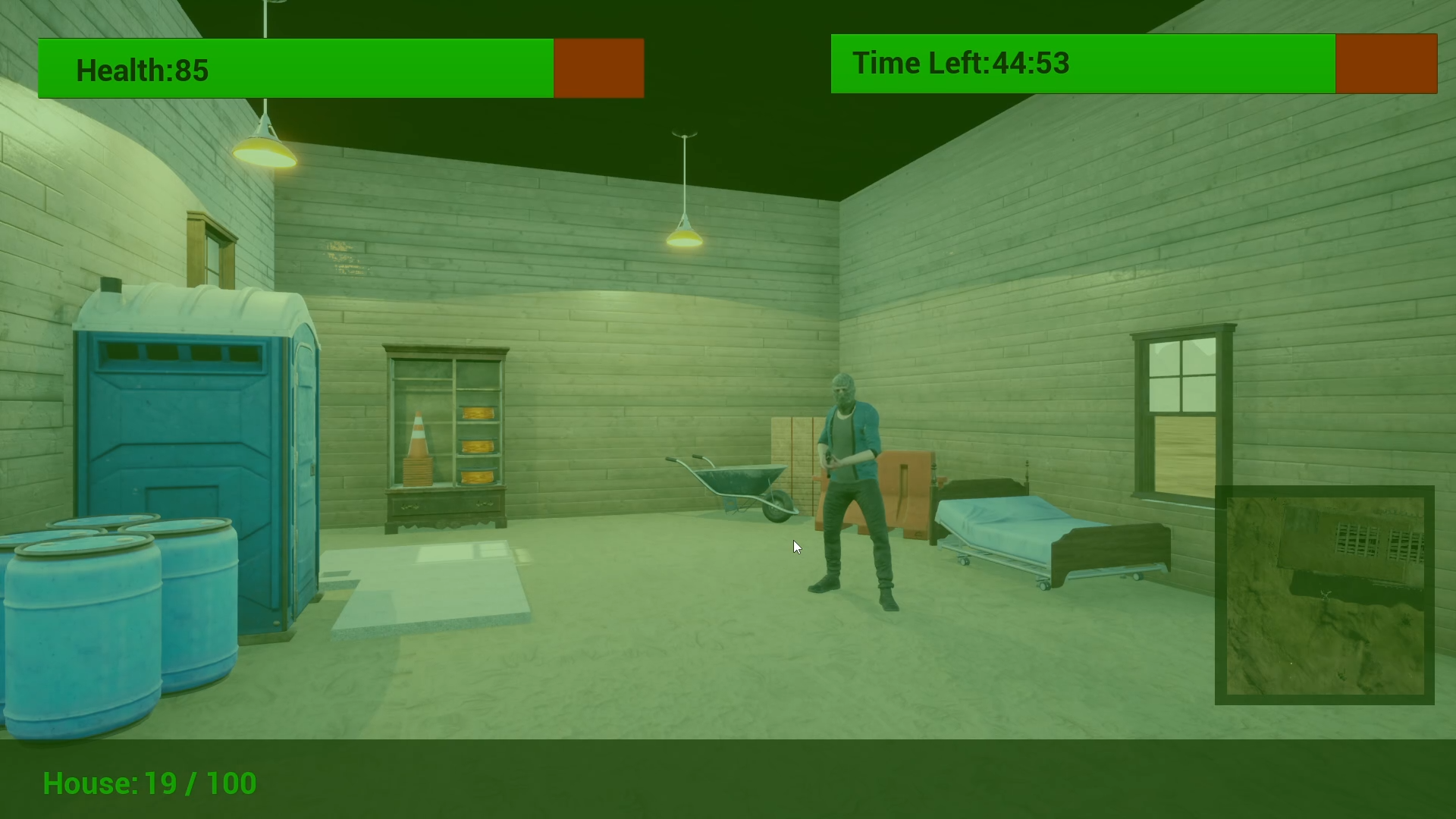}
\label{fig::Outcomes4}}
\caption{The outcomes based on threat presence and action chosen.}
\label{fig::Outcomes}
\end{figure}

We developed a 3D testbed in Unreal Engine. A screenshot of the testbed is shown in Fig. \ref{fig::Background}. Fig. \ref{fig::Recommendation} shows the recommendation dialog box where the participant was recommended to not use the RARV. After choosing an action, the four possibilities depending on the presence of threat and participant's selected actions are shown in Fig. \ref{fig::Outcomes}. The participants are told that if they encounter a threat without the RARV, they will lose $5$ points of health. They were told that deploying the RARV takes about $10$ seconds. They were told to choose an action based on their interaction history and the recommendation. Their goal was to minimize the time taken for the mission, while maintaining the soldier's health.
% After exiting each house, the participants are asked to adjust a slider to give feedback on their level of trust on the drone's recommendations.

\subsection{Measures}
\subsubsection{Personality}
The big 5 factors of personality (Extraversion, Conscientiousness, Agreeableness, Neuroticism, and Imagination) were measured using the 20-item mini-IPIP scale \cite{miniIPIPscale}. This 5-point Likert scale has widely been used in human-robot trust research \cite{miniIPIPusage}.
\subsubsection{Propensity to Trust}
Propensity to trust autonomous systems was assessed using a 6-item scale developed in \cite{MerrittPropensityScale}. This was also a 5-point Likert scale
\subsubsection{Perfect Automation Schema}
Perfect Automation Schema (PAS) was measured using the 7-item scale developed by \cite{MerrittPASscale}. Of these, 4 items measure high expectations from the autonomy and the other 3 items measure All-or-none thinking. This was a 7-point Likert scale.
\subsubsection{Moment-to-Moment Trust}
During the task, the participants were asked to rate their moment-to-moment trust on the drone by adjusting a slider on a 100-point scale. 
\subsubsection{Post-experiment trust}
After the experiment, we used two scales \cite{TrustMMScale, miniIPIPusage} for assessing trust. The first one was a 9-item questionnaire with sliders while we used 6-items from the second scale all of which were 7-point Likert type questions.
\subsubsection{Workload}
Workload was measured using the NASA Task Load Index \cite{NASATLXscale}. We only used 5 of the 6 items as there was no physical demand from the participants in our experiment. All the items had the participants rate their feelings on a slider with values ranging from Very low to Very high.

\subsection{Experimental procedure}

% \textcolor{green}{@Jessie, updated, please check}
Prior to the experiment, participants provided informed consent and completed the pre-experiment surveys. They were oriented to the steps of the experiment and walked through each of the screens they would see during the experiment. The two-fold objective of minimizing time and maximizing health was emphasized. They were informed about the performance-based bonus pay. They were told that the drone was imperfect, but they were not informed of the exact reliability level. They were also told that the robotic agent's recommendations would help them achieve the two-fold objective. Participants then proceeded to the experimental trials, wherein they had to search through 100 houses sequentially. After searching each house, the participants were asked to report their level of trust on the autonomous agent's recommendations. The participants took on average 51.5 minutes to complete the task. At the end of the experiment, participants completed the post-experiment surveys.

\section{RESULTS \& DISCUSSION}
\label{sec:result}
\subsection{Using Immediate Actual Reward as a performance metric}

\begin{figure}
    \centering
    \includegraphics[width=0.7\linewidth]{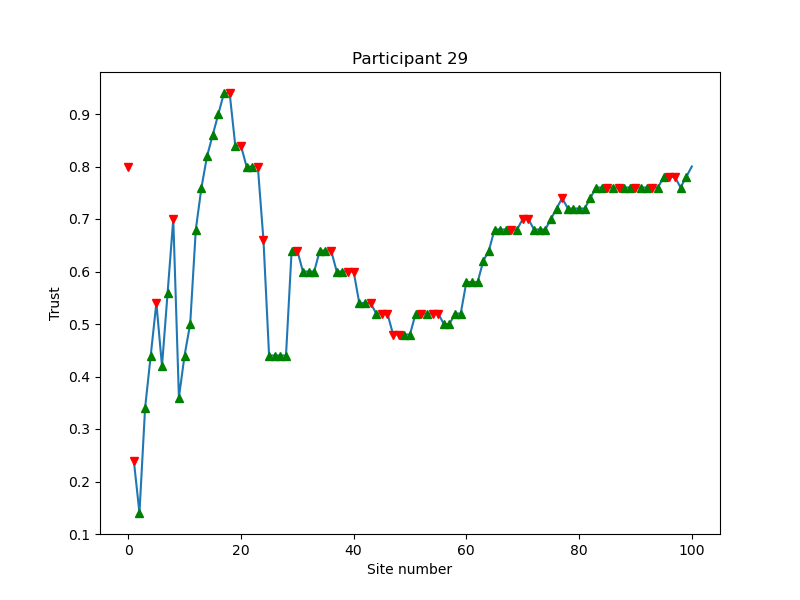}
    \caption{Trust feedback with overlay of red and green triangles representing the rewards.} 
    %Green triangles represent the participant getting a higher reward by following the recommendation and red triangles represent the opposite}
    \label{fig::ImmediateRewardAndTrust}
\end{figure}

Since the participants are explicitly told to consider both the soldier's health and the time to complete the search as their objectives, we expect their trust to be correlated with the immediate reward that they receive upon choosing an action. We expect that a participant's trust would be likely to increase if following the recommendation by the drone gets them a higher reward than doing the opposite and vice versa. Fig. \ref{fig::ImmediateRewardAndTrust} shows a representative example. In the figure, a green triangle represents the site at which following the recommendation would result in a better immediate reward gain ($p_i=1$) and the red triangles represent the opposite. It is quite clear that a red triangle is often followed by a decrease in trust and a green triangle is followed by an increase in trust. Thus, our reward-based performance metric is able to capture moment-to-moment trust changes of the participant. Using this metric in our trust update model, we get a prediction root mean squared error of $0.1266$ $(SD=0.078)$.

\subsection{Clustering of Trust Dynamics}

\begin{figure}[h]
\centering
    \includegraphics[width=0.7\linewidth]{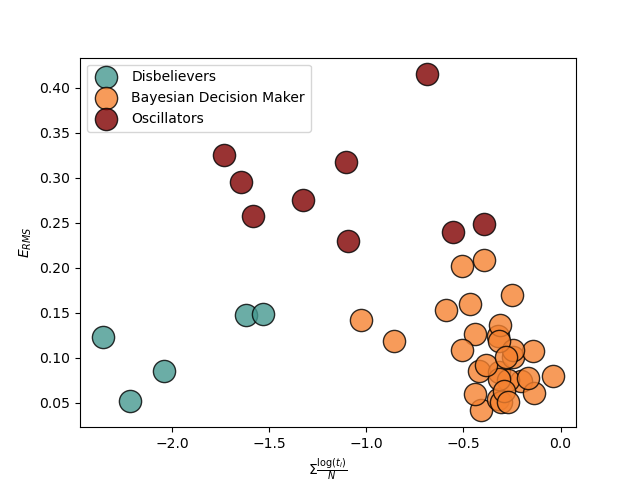}
    \caption{Clustering of participants according to their trust dynamics.} 
    % The features used are the root mean squared error between our predictions and their feedback and their average log trust.}
    \label{fig::Clustering-clusters}
\end{figure}

% \begin{figure}[h]
% \centering
% \subfigure[Variances and Silhouette Scores]{
% \includegraphics[width=2.5in]{images/Variance_Silhouette.png}
% \label{fig::Clustering-Variances}}
% \subfigure[Clusters]{
% \includegraphics[width=2.5in]{images/Clusters.png}
% \label{fig::Clustering-clusters}}
% \caption{Clustering of participants according to their trust dynamics. The features used are the root mean squared error between our predictions and their feedback and their average log trust. As evidenced by the knee in the variance plot and the maximum in the silhouette scores plot, we chose $k=3$ as the optimum number of clusters}
% \label{fig::Clustering}
% \end{figure}

We employ k-means clustering to group participants with similar trust dynamics. We use the root mean squared error ($E_{RMS}$) between the predicted trust and the trust feedback and the average logarithm of trust as the clustering features. For computing $E_{RMS}$, we consider the feedback at the first $20$ sites as a training set to fit the personalized parameters $(\alpha_0, \beta_0, w^s, w^f)$. Thereafter, we use the feedback after every 5 sites to update the parameters. Fig. \ref{fig::Clustering-clusters} shows the results. We choose $k=3$ as the optimum number of clusters because it is a turning point of both the within-cluster variance and the silhouette score. We call the cluster with small $E_{RMS}$ and generally higher values of trust \emph{Bayesian Decision Makers}. The second cluster, called \emph{Disbelievers}, consists of participants whose $E_{RMS}$ values are small but whose trust is generally low. The third significant group is the \emph{Oscillators}, whose trust changes rapidly, making it harder to predict. 

\subsection{Personal Characteristics and Type of Trust Dynamics}

% \begin{table}[h]
% \renewcommand{\arraystretch}{1.3}
% \caption{Descriptive statistics of personal characteristics between the three different trust dynamics \\(BDM = Bayesian Decision Maker)}
% \label{tab::pre-experiment}
% \centering
% \begin{tabular}{c|c|c|c}
% \hline
% \bfseries Personal Characteristic & \bfseries BDM & \bfseries Disbeliever & \bfseries Oscillator\\
% \hline
% % Age                         & 21.9 (2.7) & 24.3 (2.5) & 24.9 (3.0)\\
% Extraversion (/20) $^\ast$         & 9.5 (3.3) & 5.8 (2.8)  & 11.3 (2.9)\\
% Agreeableness (/20) $^\ast$      & 13.5 (2.5) & 10.4 (5.0) & 14.1 (1.8)\\
% Conscientiousness (/20)     & 13.1 (2.7) & 12.4 (3.0) & 12.1 (4.5)\\
% Neuroticism (/20)           & 7.9 (2.7)  & 6.8 (3.6)  & 10.2 (4.7) \\
% Intellect/Imagination (/20) $^\dagger$  & 11.7 (2.0) & 9.8 (1.8)  & 12.2 (1.8)\\
% High Expectations (/28) $^{\ast\ast}$    & 12.7 (3.9) & 6.4 (2.8)  & 12.4 (4.2)\\
% All or None Thinking (/21)  & 6.6 (2.9)  & 6.4 (3.4)  & 7.1 (3.1 ) \\
% Trust Propensity (/30) $^\dagger$     & 20.2 (4.4) & 17.2 (4.1) & 22.8 (3.2)\\
% \hline
% \end{tabular}
% \end{table}

We performed one-way ANOVA between the data of the clustered participants. Results showed significant difference between the three types of trust dynamics in Extraversion ($F(2,42)=4.991, p=0.011$), Agreeableness ($F(2,42)=3.276, p=0.048)$, and the High Expectations facet of the Perfect Automation Schema $(F(2,42)=5.752, p=0.006)$. Further, propensity to trust automation $(F(2,42)=3.002, p=0.06)$ and intellect/imagination $(F(2,42)=2.687,p=0.08)$ had marginally significant differences. 

% \begin{figure*}[ht]
% \centerline{
% \subfigure[Extraversion]{
% \includegraphics[width=2.1in]{images/BarCharts/Extraversion.png}
% \label{fig::extraversion}}
% \hfil
% \subfigure[Agreeableness]{
% \includegraphics[width=2.1in]{images/BarCharts/Agreeableness.png}
% \label{fig::agreeableness}}
% \hfil
% \subfigure[High Expectations]{
% \includegraphics[width=2.1in]{images/BarCharts/HighExpectations.png}
% \label{fig::he}}
% }
% \caption{Association between personal characteristics and type of trust dynamics.\\\hspace{\textwidth}
% \small{** $p<0.01$; * $p<0.05$; $\dagger$ $p<0.1$ }, BDM=Bayesian Decision Maker}
% \label{fig::post-hoc-results}
% \end{figure*}

% \begin{table}[h]
% \renewcommand{\arraystretch}{1.3}
% \caption{Descriptive statistics of post experiment metrics between the three different trust dynamics}
% \label{tab::post-experiment}
% \centering
% \begin{tabular}{c|c|c|c}
% \hline
% \bfseries Personal Characteristic & \bfseries BDM & \bfseries Disbeliever & \bfseries Oscillator\\
% \hline
% Trust (Muir) (/100) $^{\ast\ast\ast}$     & 65.4 (13.5) & 15.8 (9.9) & 44.7 (26.1)\\
% Trust (Lyons) (/7)  $^{\ast\ast\ast}$    & 4.5 (0.54)  & 3.1 (0.6)   & 3.6 (0.9)\\
% Mental Demand (/100)    & 39.6 (25.2) & 42.0 (36.6) & 50.3 (28.6)\\
% Temporal Demand (/100)  & 50.8 (27.4) & 62.0 (24.5) & 42.9 (21.3)\\
% Performance (/100)      & 58.6 (19.7) & 50.8 (30.0) & 46.2 (31.7)\\
% Effort (/100)           & 34.3 (23.0) & 34.4 (17.4) & 49.8 (32.2)\\
% Frustration (/100) $^{\ast}$      & 45.8 (22.2) & 58.4 (25.4) & 68.1 (14.3)\\
% \hline
% \end{tabular}
% \end{table}
Post-hoc analysis with Bonferroni adjustment shows that the disbelievers are significantly less extroverted than the oscillators $(p=0.009)$. Disbelievers were only marginally less extroverted than Bayesian Decision Makers $(p=0.061)$. Disbelievers were marginally less agreeable than the Bayesian Decision Makers ($p=0.068$)  and Oscillators ($p=0.06$). Disbelievers had significantly lower expectations from automation compared to both Bayesian Decision Makers $(p=0.005)$ and Oscillators $(p=0.023)$.
%Fig. \ref{fig::post-hoc-results} summarizes these findings.

One way ANOVA on the post-experiment measures show significant difference between the three clusters in their post-experiment trust reports (Trust questionnaire by Muir and Moray $F(2,42)=22.167, p<0.001$, Trust questionnaire by Lyons and Guznov $F(2,42)=15.183, p<0.001$) and their frustration levels ($F(2,42)=4.136, p=0.023)$).

Post-hoc analysis with Bonferroni adjustment shows that there are significant differences between each of the three groups' trust reports according to the trust questionnaire by Muir and Moray ($p<0.001$ between Bayesian Decision Makers and Disbelievers, $p=0.006$ between Bayesian Decision Makers and Oscillators, and $p=0.009$ between Oscillators and Disbelievers with the highest trust for Bayesian Decision Makers and lowest for Disbelievers). Trust reported with the questionnaire by Lyons and Guznov only showed significant difference between Disbelievers and Bayesian Decision Makers $(p<0.001)$, and between Oscillators and Bayesian Decision Makers $(p=0.002)$. Oscillators were significantly more frustrated than Bayesian Decision Makers $(p=0.025)$.

\section{CONCLUSIONS}
\label{sec:discussion}
In this study, we formulated the human-robot sequential decision-making problem as an MDP and proposed an innovative reward-based performance metric for trust-aware decision-making. Through a human-subject experiment, we found three distinct types of trust dynamics, namely, Bayesian Decision Makers, Oscillators, and Disbelievers. Our model achieved different accuracies on different types on trust dynamics, which points to a requirement to use different models for people in different categories.

Analyses suggest that those individuals classified as Bayesian Decision makers evidenced high expectations of automation. When combined with other state measures such as trust and frustration, one might be able to identify a profile of the Bayesian Decision makers as the combination of individual differences and state measures could be used to parse the sample into the three categories of trust dynamics. Knowing that an individual might fall into one of the categories could influence whether or not a machine partner that is equipped with a dynamic trust model is a feasible solution for that individual.      

In future studies, besides the reverse-psychology model, different human trust-behavior models should be considered. In addition, techniques such as inverse reinforcement learning can be employed to learn the preference of the human teammate to improve the team performance.

\bibliographystyle{IEEEtran}
\bibliography{IEEEabrv, mybibfile}

\end{document}